\documentclass{article}

\usepackage[preprint]{neurips_2026}

\usepackage[utf8]{inputenc}
\usepackage[T1]{fontenc}
\usepackage{hyperref}
\usepackage{url}
\usepackage{booktabs}
\usepackage{amsfonts}
\usepackage{amsmath}
\usepackage{amssymb}
\usepackage{nicefrac}
\usepackage{microtype}
\usepackage{xcolor}
\usepackage{colortbl}
\definecolor{sharpblue}{HTML}{4A6FA5}
\newcommand{\rA}[1]{\cellcolor{sharpblue!26}\textbf{#1}}
\newcommand{\rB}[1]{\cellcolor{sharpblue!18}\underline{#1}}
\newcommand{\rC}[1]{\cellcolor{sharpblue!13}#1}
\newcommand{\rD}[1]{\cellcolor{sharpblue!9}#1}
\newcommand{\rE}[1]{\cellcolor{sharpblue!6}#1}
\newcommand{\rF}[1]{\cellcolor{sharpblue!3}#1}
\newcommand{\rG}[1]{\cellcolor{sharpblue!1}#1}
\usepackage{graphicx}
\usepackage{multirow}
\usepackage[capitalize,noabbrev]{cleveref}
\usepackage{enumitem}
\usepackage{algorithm}
\usepackage{algorithmic}
\usepackage{subcaption}
\usepackage{wrapfig}
\usepackage{xspace}
\usepackage{tikz}
\usetikzlibrary{arrows.meta,positioning,fit,backgrounds,calc,decorations.pathreplacing}
\usepackage[most]{tcolorbox}

\newtcolorbox{seedrule}[1][]{
    enhanced,
    boxrule=0.4pt,
    colback=gray!4,
    colframe=gray!40,
    arc=3pt,
    left=4pt, right=4pt, top=3pt, bottom=3pt,
    boxsep=0pt,
    fontupper=\small,
    #1
}

\newtcolorbox{attrstep}[1]{
    enhanced,
    boxrule=0.4pt,
    colback=blue!2,
    colframe=blue!20,
    arc=3pt,
    left=6pt, right=6pt, top=4pt, bottom=4pt,
    boxsep=0pt,
    fontupper=\small,
    title={#1},
    coltitle=black!80,
    fonttitle=\bfseries\small,
    colbacktitle=blue!5,
    attach boxed title to top left={yshift=-2mm, xshift=4mm},
    boxed title style={boxrule=0.4pt, colframe=blue!20, arc=2pt},
}

\usepackage{fancyvrb}
\definecolor{diffgreentext}{HTML}{116329}
\definecolor{diffredtext}{HTML}{82071E}
\definecolor{diffgray}{HTML}{656D76}
\newcommand{\method}{SHARP\xspace}

\usepackage{pifont}

\title{SHARP: A Self-Evolving Human-Auditable Rubric Policy for Financial Trading Agents}

\author{
  \textbf{Xiwen Chen}\textsuperscript{$\dagger$,}\thanks{Equal contribution.} \quad
  \textbf{Wenhui Zhu}\textsuperscript{$\ddagger$,}\footnotemark[1] \quad
  \textbf{Songzhu Zheng}\textsuperscript{$\dagger$,}\footnotemark[1] \\
  \textbf{Kashif Rasul}\textsuperscript{$\dagger$} \quad
  \textbf{Yueyue Deng}\textsuperscript{$\S$} \quad
  \textbf{Huayu Li}\textsuperscript{$\P$} \\
  \textsuperscript{$\dagger$}Morgan Stanley \quad
  \textsuperscript{$\ddagger$}Arizona State University \\
  \textsuperscript{$\S$}Columbia University \quad
  \textsuperscript{$\P$}University of Arizona
}
\begin{document}

\maketitle

\begin{abstract}
Large language models (LLMs) are increasingly deployed for autonomous financial trading, a domain requiring continuous adaptation to noisy, non-stationary markets. Existing self-improving agents typically address this through unbounded free-form prompt optimization. However, in low signal-to-noise environments with delayed scalar rewards (P\&L), this unstructured approach exacerbates the fundamental credit assignment problem: optimizers cannot reliably distinguish systematic logic flaws from stochastic market variance, inevitably leading to policy drift. 
To overcome this bottleneck, we introduce the Self-Evolving Human-Auditable Rubric Policy (\method), a neuro-symbolic framework that replaces unconstrained text mutation with structured, symbolic policy optimization. \method confines the agent's reasoning to a bounded, human-readable rubric of explicit condition-action rules. When sub-optimal trades occur, an attribution agent employs cross-sample reasoning across multiple samples to isolate specific rule failures. This enables targeted, atomic policy edits that are subsequently regularized through strict walk-forward validation. 
Evaluated across three diverse equity sectors and four LLM backbones, \method consistently transforms generic initial heuristics into highly robust strategies, lifting the empirical performance of compact models by 10 to 20 percentage points on average (e.g., GPT-4o-mini). Ultimately, \method demonstrates that LLMs can achieve dynamic and efficient adaptation while significantly enhancing the structural transparency and auditability demanded by institutional finance\footnote{\textcolor{red}{Opinions expressed in this paper are those of the authors, and do not necessarily reflect the view of Morgan Stanley (See \textbf{Disclaimer} section).}}.
\end{abstract}
\section{Introduction}
\label{sec:intro}

Large language models (LLMs) have demonstrated remarkable prowess in complex reasoning tasks, frequently rivaling human performance in mathematical deduction, code synthesis, and scientific heuristics~\citep{openai2024gpt4o, deepseek2025r1}. This has catalyzed a surge of interest in deploying LLMs for autonomous financial decision-making. Recent architectures have evolved from simple news classifiers~\citep{lopezlira2023chatgpt} and retrieval-augmented agents~\citep{fatouros2025marketsenseai} to sophisticated multi-agent systems featuring hierarchical memory~\citep{yu2024finmem} and self-critiquing belief systems~\citep{yu2024fincon}. These developments underscore a fundamental consensus: financial signals are inherently noisy and markets are non-stationary, so a trading agent must possess the capacity for robust, continuous self-adaptation against low signal-to-noise ratio (SNR) environments.

We agree with this premise, but take a different view on \emph{how} adaptation should work.
Trading is a feedback-driven process in which every decision is scored against the market, and the goal is not to be clever but to be systematically less wrong over time.
When quantitative traders and researchers lose money on a trade, they do not throw away their strategy and start over.
They perform structural credit assignment and isolate the specific faulty heuristic, perhaps they over-weighted a news item that was already priced into the stock, adjust that rule, and leave the rest of the playbook intact.
This is how disciplined quantitative research actually works: \emph{diagnose before you prescribe}. This disciplined process represents a powerful inductive bias that remains absent from current LLM-based systems.

Current LLM-based trading agents struggle to balance this dynamic adaptation with structural rigor. Adaptation in prior work, whether via layered memory~\citep{yu2024finmem}, investment beliefs~\citep{yu2024fincon}, or prompt optimization~\citep{yang2023large, agrawal2025gepa}, updates unstructured natural-language state through free-form text edits. Recent pioneering frameworks like ATLAS~\citep{papadakis2025atlas} demonstrate the necessity of adapting to non-stationary markets through dynamic prompt optimization. However, we argue that reliance on \emph{unbounded free-form text mutation} faces a fundamental optimization bottleneck. 
\begin{wrapfigure}{r}{0.55\textwidth} 
  \vspace{-2pt} 
  \centering
  \includegraphics[width=\linewidth]{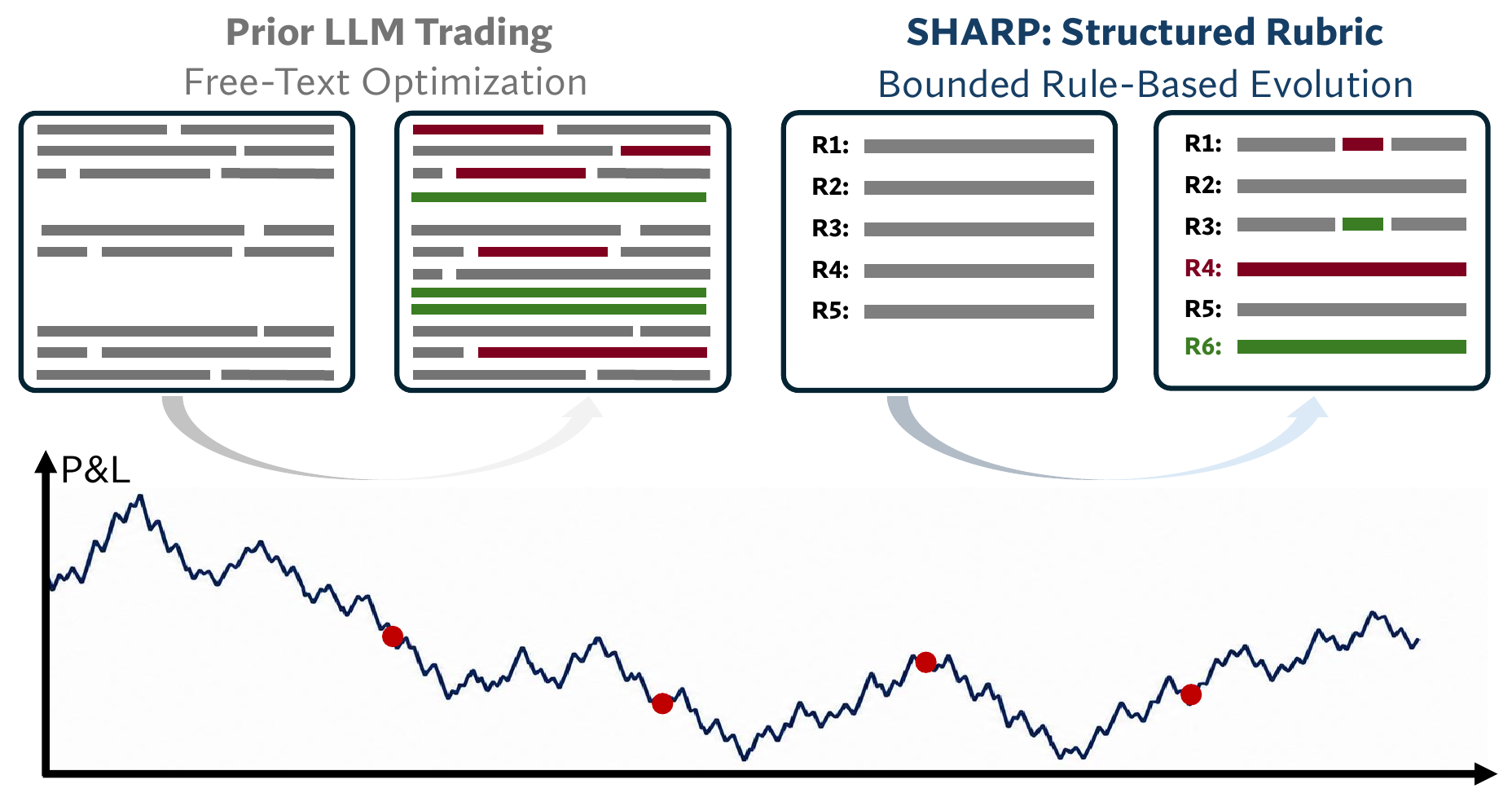}
  \caption{\textbf{Overcoming the Credit Assignment Problem in LLM-Based Trading.} (\textbf{Left}) Standard self-improving agents rely on unbounded free-text optimization. When facing noisy P\&L feedback, they cannot isolate logical errors, leading to untargeted mutations and policy degeneration. (\textbf{Right}) SHARP introduces a structured, human-auditable rubric. Losses are symbolically attributed to specific rules, enabling targeted, atomic edits. }
  \label{fig:teaser}
  \vspace{-10pt} 
\end{wrapfigure}
First, the \emph{search space is practically intractable}: in a free-form prompt, any token may change at any step, creating an unbounded and high-dimensional discrete search space. When coupled with the inherently noisy and non-stationary reward signals (P\&L) of financial markets, finding a stable optimal prompt becomes prohibitively difficult, as the optimizer cannot reliably distinguish structural systematic signal from stochastic variance, inevitably leading to variance-induced policy drift. Second, unstructured text mutations suffer from a severe \emph{credit assignment problem}~\citep{sutton1998reinforcement}: although the optimizer receives aggregate performance feedback, it lacks a mechanism to attribute a specific portfolio loss to a localized logical error. Without structural boundaries, the model cannot isolate which part of the reasoning failed, leading to untargeted rewrites rather than precision corrections. We reproduce this failure mode empirically (\cref{sec:ablation}): replacing the structured rubric with free-form reflection turns a strongly positive AI~Tech return into a sizeable negative one (from roughly $+33\%$ to $-12\%$), localizing the failure to the lack of structural constraints in the mutation operator.

We propose \textit{a Self-Evolving Human-Auditable Rubric Policy} (\method, see \cref{fig:teaser}), a neuro-symbolic framework that closes this gap by replacing unconstrained prompt mutation with structured, symbolic policy optimization. This method reintroduces the two properties that make quantitative research work: \emph{structure} and \emph{attribution}. Instead of a free-form prompt, the LLM reasons under a \emph{rubric}, a bounded set of condition-action rules such as ``\textit{IF VIX is above 25, THEN reduce bullish return estimates by 30\%.}''
An \emph{attribution agent} performs retrospective diagnosis to trace P\&L failures back to specific Rule IDs, disentangling systematic flaws from market noise.
An \emph{evolution agent} then executes targeted, atomic rule edits within a bounded symbolic space: modify one rule's threshold, add a new rule for an unhandled pattern, or remove a rule that has been misfiring.
A \emph{validation gate} accepts the edit only if it improves held-out performance, preventing overfitting to recent noise.
Every rule, every edit, and every attribution trace is human-readable, versioned, and auditable.

This design draws inspiration from recent work on rubric mining in LLM alignment~\citep{xie2025autorubric}: Auto-Rubric mines interpretable rubrics from preference data, while \method evolves trading rubrics from live P\&L feedback through an iterative attribution, mutation, and validation loop.
Our framework is governed by two core structural constraints.
(i)~The rubric constrains \emph{what} can evolve (structured condition/action pairs, not arbitrary text) and \emph{how much} can evolve (a bounded number of rule edits per round). This turns an intractable text search into a tractable structured one, a highly efficient approach that effectively leverages the reasoning potential of compact models without relying solely on massive compute scaling.
(ii)~The financial judgment resides in the rubric and is refined from market feedback, while the LLM's role is confined to mapping unstructured news into the rubric's condition space, a narrower and more inspectable task than free-form return reasoning.

This work poses a fundamental question distinct from prior LLM trading systems: \emph{can an autonomous agent dynamically adapt to noisy, non-stationary environments without sacrificing the strict logical transparency required by quantitative finance?} We present empirical evidence that the answer is yes. Our main contributions are threefold:
\begin{itemize}
    \item \textbf{Neuro-Symbolic Rubric Learning Paradigm.} We introduce \emph{rubric evolution} as a principled alternative to free-form prompt optimization. By formalizing LLM-based trading as a constrained optimization over symbolic condition-action sets, \method effectively mitigates variance-induced policy drift and ensures that the strategy search space remains tractable.
    
    \item \textbf{Attribution-Guided Credit Assignment.} We design a tri-agent evolution loop (Attribution, Evolution, and Validation) that tackles the severe credit assignment problem inherent in financial markets. By employing structural credit assignment to diagnose localized rule-level failures before proposing atomic edits, \method effectively separates systematic logic flaws from stochastic market noise.
    
    \item \textbf{Empirical Generalization and Auditability.} Evaluated under a rigorous walk-forward protocol across three diverse equity sectors (AI Tech, Biotech, Consumer Discretionary) and four LLM backbones, \method consistently transforms generic initial heuristics into highly competitive strategies against established baselines, while maintaining a fully transparent, versioned, and human-auditable policy trace.
\end{itemize}

\section{Related Work}
\label{sec:related}

\noindent\textbf{LLM-based trading and self-improving agents.}
Early applications of LLMs in finance demonstrated that sentiment analysis of news headlines can potentially predict market movements~\citep{lopezlira2023chatgpt}, leading to sophisticated multi-agent architectures like MarketSenseAI~2.0~\citep{fatouros2025marketsenseai}. However, these systems rely on fixed prompts, leaving them vulnerable to market drift. A subsequent wave of self-improving agents attempts to address this via free-form state updates: \citep{yu2024finmem,li2023tradinggpt,zhang2024multimodal} utilize layered memory, FinCon~\citep{yu2024fincon} employs a multi-agent manager-analyst hierarchy that updates free-form investment beliefs via conceptual verbal reinforcement, and MountainLion~\citep{wu2025mountainlion} integrates a free-form reflection module over multimodal inputs. While these architectures introduce adaptation and multi-agent collaboration, they share a structural limitation: adaptation occurs through unconstrained natural-language updates to an unstructured state, making the internal logic hard to audit. Parallel efforts in prompt optimization, such as ATLAS~\citep{papadakis2025atlas}, have pioneered the application of OPRO~\citep{yang2023large} to dynamic trading. Yet, similar to general-purpose optimizers like \citep{agrawal2025gepa} and \citep{opsahl2024miprov2}, these methods treat the prompt as a monolithic string. In noisy financial environments, updating this global state without precise fault isolation leads to the \emph{credit assignment problem} discussed in \cref{sec:intro}. \method addresses this by shifting the paradigm: (i)~an attribution agent logically isolates losses to specific rules, ensuring targeted rather than global updates, and (ii)~the search space is strictly bounded to structured, interpretable rule edits.


\noindent\textbf{Signal mining and automated ML.}
Parallel work uses LLMs to discover quantitative signals~\citep{han2026quantaalpha, tang2025alphaagent, li2025rdagentq} or automate the broader ML research cycle~\citep{li2025rdagent, jiang2025aide, hogan2026alphalab}. 
Crucially, these methods discover \emph{numerical} formulas or \emph{models} from structured data; \method instead evolves \emph{natural-language} analysis rules for unstructured text reasoning, making the paradigms orthogonal and complementary.

\noindent\textbf{Rubric mining and symbolic credit assignment.}
In the alignment literature, Reinforcement Learning with Verifiable Rewards (RLVR), such as GRPO~\citep{shao2024deepseekmath}, relies on strict, deterministic reward functions (e.g., mathematical correctness). Such verifiable signals are fundamentally unavailable in financial trading, where rewards (P\&L) are delayed, stochastic, and subject to extreme noise. To provide structure in open-ended tasks where rewards are either non-verifiable or black-boxes, explicit rubrics have emerged as a solution. Frameworks like \citep{gunjal2025rubrics,xie2025autorubric,rrd2026,huang2025reinforcement} demonstrate that mining fine-grained criteria from \emph{static preference data} can effectively transform black-box reward models into interpretable evaluation tools.

While \method shares this structural insight, it fundamentally repurposes the rubric paradigm for a non-stationary setting. Unlike Rubric as Reward Function, which extracts criteria from offline datasets to improve \emph{passive evaluation}, \method \emph{evolves} the rubric dynamically from live P\&L to guide \emph{active decision-making}. By doing so, the rubric acts as a \emph{symbolic credit assignment} mechanism: it maps noisy, scalar P\&L signals logically back to discrete rule IDs. This provides a tractable alternative to policy gradients or deterministic RLVR environments, enabling systematic strategy refinement even in the presence of intense market noise.
\section{Method}
\label{sec:method}

\subsection{Problem Formulation}
\label{sec:problem}

We consider a daily long-short equity trading problem over a universe of $N$ stocks $\mathcal{U} = \{u_1, \ldots, u_N\}$.
On each trading day $t$, the decision-maker observes an information set
$
    \mathcal{I}_t = \bigl(\,\mathcal{N}_t,\; \mathbf{P}_t,\; \mathbf{m}_t\,\bigr)
$
consisting of news articles $\mathcal{N}_t = \{\mathcal{N}_t^{(i)}\}_{i=1}^N$ per stock (timestamp-filtered to prevent lookahead), price features $\mathbf{P}_t = \{\mathbf{p}_t^{(i)}\}_{i=1}^N$ (e.g., recent returns, 52-week range), and macro context $\mathbf{m}_t$ (i.e., market trend, volatility, rates).
An LLM analyst $f_\theta$ maps each stock's information to a trading signal together with an activation trace:
\begin{equation}
    f_\theta\bigl(\mathcal{N}_t^{(i)},\, \mathbf{p}_t^{(i)},\, \mathbf{m}_t \;\big|\; \mathcal{R}\bigr) \;\longmapsto\; \bigl(\hat{r}_t^{(i)},\, c_t^{(i)},\, \mathcal{A}_t^{(i)}\bigr),
\end{equation}
where $\hat{r}_t^{(i)}$ is the predicted return, $c_t^{(i)} \in [0,1]$ is the confidence, $\mathcal{A}_t^{(i)} \subseteq \{1,\ldots,M\}$ is the set of activated rule indices, and $\mathcal{R}$ is a set of analysis instructions.
The composite score $\sigma_t^{(i)} = \hat{r}_t^{(i)} \cdot c_t^{(i)}$ determines the portfolio: long the top-$K$ stocks, short the bottom-$K$.
Because $f_\theta$ processes each stock independently, all cross-sectional operations (ranking by $\sigma$, portfolio construction) occur outside the LLM in a deterministic aggregation step.


The central question is: \emph{what should $\mathcal{R}$ be, and how should it adapt?}
Prior work typically treats $\mathcal{R}$ as free-form prompt text. While recent methods have demonstrated the necessity of updating this prompt dynamically, optimizing a monolithic string under noisy financial feedback introduces a severe \textbf{credit assignment problem}: there is no formal mechanism to isolate and attribute a portfolio drawdown to a specific logical error. 
To resolve this, we propose a fundamentally different approach: \textbf{\textit{$\mathcal{R}$ is not free-form text, but a neuro-symbolic rubric, establishing a tractable, discrete search space for policy optimization.}} By defining $\mathcal{R}$ as a bounded set of structured condition-action rules ($|\mathcal{R}| \leq M_{\max}$), updates are mathematically restricted to atomic policy edits rather than global text rewrites. We validate the necessity of these structural constraints empirically in \cref{sec:ablation}.

\subsection{Rubric: A Bounded Set of Structured Rules}
\label{sec:rubric}

A rubric $\mathcal{R} = \{R_1, \ldots, R_M\}$ is a versioned set of analysis rules, where each rule is a structured tuple:
\begin{equation}
    R_k = \bigl(\texttt{id}_k,\;\texttt{cat}_k,\;\texttt{cond}_k,\;\texttt{act}_k\bigr).
    \label{eq:rule}
\end{equation}
Here $\texttt{id}_k$ is a unique identifier (e.g., \texttt{news\_fda\_action}); $\texttt{cat}_k$ is a semantic category that groups rules by function; $\texttt{cond}_k$ is a natural-language predicate specifying when the rule fires (e.g., ``FDA approval mentioned in news''); and $\texttt{act}_k$ is the prescribed adjustment to the LLM's analysis (e.g., ``increase signal weight by $2.5\times$'').
Categories organize rules by function.
We illustrate each with a representative initial rule:

\begin{seedrule}
\textbf{\emph{Temporal discounting}} (adjusts for moves already reflected in price).\\
\textbf{IF} the stock moved ${>}3\%$ in the news direction over the past 5 days \textbf{THEN} reduce signal strength by 70\%.
\end{seedrule}

\begin{seedrule}
\textbf{\emph{News weighting}} (scales the signal by event type).\\
\textbf{IF} news mentions an FDA approval or rejection \textbf{THEN} treat as very high impact: weight ${\times}\,2.5$.
\end{seedrule}

\begin{seedrule}
\textbf{\emph{Macro interaction}} (conditions on market-wide indicators).\\
\textbf{IF} VIX is above 25 \textbf{THEN} reduce all bullish return estimates by 30\%.
\end{seedrule}

\begin{seedrule}
\textbf{\emph{News scarcity}} (reduces confidence under limited information).\\
\textbf{IF} fewer than 3 news articles are available for the ticker \textbf{THEN} cap confidence at 0.4.
\end{seedrule}

The rubric is injected into the LLM's context as a structured symbolic block. The LLM is constrained to report the activated rule IDs in its output, producing the trace $\mathcal{A}_t^{(i)}$ in Eq.~1. Because rule adjustments are executed by the LLM's own generation rather than a programmatic engine, $\mathcal{A}$ is a \emph{self-reported attribution} rather than a formally deterministic trace; nevertheless, it provides a tractable credit assignment mechanism absent in free-form prompt optimization.
We show empirically (\cref{sec:ablation}; Appendix~\ref{app:reflect_example}) that even when initialized with identical rubric knowledge, free-form reflection gradually dissolves the structured rules into untraceable prose, whereas \method's atomic edits preserve a full audit trail from each threshold change back to the error days that motivated it.

\subsection{Rubric Evolution}
\label{sec:evolution}

\begin{figure}[t]
\centering
\resizebox{\linewidth}{!}{%
\begin{tikzpicture}[
    node distance=10mm and 12mm,
    font=\small,
    base/.style={draw=black!50, thick, rounded corners=4pt, minimum height=12mm, minimum width=28mm, align=center, fill=white, text=black!90},
    llm/.style={base, fill=blue!4, draw=blue!40, thick},
    data/.style={base, fill=gray!8, draw=gray!40, rounded corners=2pt},
    arr/.style={->, >=stealth, thick, draw=black!60},
    lbl/.style={font=\footnotesize, text=black!70},
    phase/.style={draw=black!20, dashed, rounded corners=6pt, inner sep=12pt, fill=black!2},
]

\node[data] (rubric) {Current Rubric\\$\mathcal{R}^{(j)}$};
\node[llm, right=10mm of rubric] (attr) {Attribution Agent\\(Backtest \& Diagnose)};
\node[llm, right=12mm of attr] (evo) {Evolution Agent\\(Mutate Rules)};
\node[base, right=15mm of evo] (val) {Validation Gate\\(OOS Test)};

\draw[arr] (rubric) -- (attr) node[lbl, midway, above] {$\mathcal{D}_{\text{train}}$};
\draw[arr] (attr) -- (evo) node[lbl, midway, above] {errors $\mathcal{E}$};
\draw[arr] (evo) -- (val) node[lbl, midway, above] {candidate $\tilde{\mathcal{R}}$};

\coordinate (loop1) at ($(val.south) + (0,-0.6)$);
\coordinate (loop2) at ($(rubric.south) + (0,-0.6)$);
\draw[arr] (val.south) -- (loop1) -- (loop2) node[lbl, midway, above] {Accept if better, else Reject} -- (rubric.south);

\begin{scope}[on background layer]
\node[phase, fit=(rubric)(attr)(evo)(val)(loop1)(loop2),
      label={[font=\small\bfseries, text=black!80, anchor=north west, inner sep=4pt]north west:{Training Phase (Iterative Rubric Evolution)}}]
      (trainbox) {};
\end{scope}

\node[data, below=18mm of rubric] (input) {Daily Market Data\\$(\mathcal{N}_t,\;\mathbf{P}_t,\;\mathbf{m}_t)$};
\node[llm, below=18mm of attr] (llm) {LLM Analyst $f_\theta(\cdot)$\\w/ Frozen Rubric $\mathcal{R}^*$};
\node[base, below=18mm of evo] (port) {Trading Signals\\$(\hat{r}_i, c_i, \mathcal{A}_i)$};
\node[data, below=18mm of val] (exec) {Portfolio Execution\\(5L / 5S, O2O Returns)};

\draw[arr] (input.east) -- (llm.west);
\draw[arr] (llm.east) -- (port.west) node[lbl, midway, above] {rank $\sigma_i$};
\draw[arr] (port.east) -- (exec.west);

\begin{scope}[on background layer]
\node[phase, fit=(input)(llm)(port)(exec),
      label={[font=\small\bfseries, text=black!80, anchor=north west, inner sep=4pt]north west:{Inference Phase (Walk-forward Out-of-Sample)}}]
      (infbox) {};
\end{scope}

\end{tikzpicture}%
}
\caption{\method pipeline.
\textbf{Top}: training-phase evolution loop.
Each round backtests the current rubric, attributes losses to rules, proposes mutations, and validates before accepting.
\textbf{Bottom}: at inference, the frozen rubric $\mathcal{R}^*$ and daily market data feed into the LLM analyst to produce trading signals.}
\label{fig:pipeline}
\end{figure}
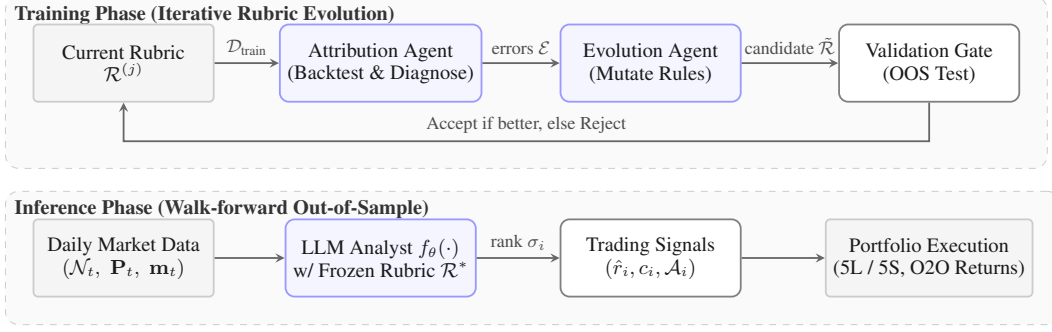

\cref{fig:pipeline} illustrates the two phases of \method.
During \emph{training}, the rubric is iteratively refined through a three-agent loop; during \emph{inference}, the best rubric is frozen and the LLM analyst produces daily trading signals.

\noindent\textbf{Training: three-agent evolution loop.}
Given training data $\mathcal{D}_{\text{train}}$ and held-out validation data $\mathcal{D}_{\text{val}}$, the rubric evolves over $J$ rounds (see Appendix~\ref{app:algorithm}).
Each round proceeds in three phases:

\ding{192}~\textit{Attribution (Symbolic Credit Assignment).}
The current rubric $\mathcal{R}^{(j-1)}$ is backtested on $\mathcal{D}_{\text{train}}$, executing a full inference loop to generate positions and P\&L. An \emph{attribution agent} then examines the $K_{\text{attr}}$ worst-performing portfolio days to perform structural credit assignment. This selective focus on the left tail of the P\&L distribution ensures that the evolution prioritizes the mitigation of severe drawdowns over average predictive accuracy, effectively acting as an automated tail-risk truncation mechanism. Rather than merely parsing logs, it conducts cross-sample diagnosis: identifying \emph{systematic} logic flaws from stochastic market noise by asking which rule adjustments would have prevented the drawdown. Sporadic variance is filtered out, ensuring only recurring failure modes drive policy updates.
\cref{app:attribution} provides a complete trace of this attribution process. For each of the $K_{\text{attr}}$ worst days, the agent analyzes the full state: asset holdings, predicted returns, activated rules, realized P\&L, and news context. The agent synthesizes this to output a structured error pattern, mapping the portfolio loss to a specific rubric vulnerability (e.g., a miscalibrated threshold in \texttt{temporal\_priced\_in}).

\ding{193}~\textit{Evolution (Symbolic Mutation).}
An \emph{evolution agent} receives the current rubric and the attributed error patterns $\mathcal{E}$. Instead of generating unbounded text, it executes a constrained search, proposing a set of \emph{atomic mutations} $\mathcal{P}$ within the symbolic space. These mutations, synthesizing new rules, calibrating thresholds ($\texttt{cond}_k$), or adjusting signal weights ($\texttt{act}_k$), are explicitly linked to the causal diagnostics provided by the attribution step.

\ding{194}~\textit{Validation Gate (Empirical Regularization).}
The candidate rubric $\tilde{\mathcal{R}}^{(j)}$ is backtested on the held-out $\mathcal{D}_{\text{val}}$. It is accepted if its excess return satisfies
$
    e\bigl(\tilde{\mathcal{R}}^{(j)}\bigr) \;\geq\; e\bigl(\mathcal{R}^*\bigr) - \epsilon,
    \label{eq:validation}
$
where $e(\cdot)$ is the validation excess return and $\mathcal{R}^*$ is the best rubric seen so far. The tolerance $\epsilon$ acts as a hyperparameter for exploration, preventing the system from overfitting to the immediate training distribution. This walk-forward gate serves as a critical regularizer against variance-induced policy drift. Rejected candidates are discarded and the previous rubric carries forward.

All hyperparameters (evolution budget $J$, validation tolerance $\epsilon$, attribution window $K_{\text{attr}}$, mutation bounds, and attribution thresholds) are specified in \cref{sec:setup}.

\noindent\textbf{Inference.}
After training, the best-validated rubric $\mathcal{R}^*$ is frozen.
On each test day $t$, the LLM analyst $f_\theta$ receives the day's information $\mathcal{I}_t$ together with $\mathcal{R}^*$, produces per-stock signals $(\hat{r}_t^{(i)}, c_t^{(i)})$, and the portfolio is constructed by ranking $\sigma_t^{(i)} = \hat{r}_t^{(i)} \cdot c_t^{(i)}$.
No rubric changes occur during test.

\vspace{-0.1in}
\section{Experiments}
\label{sec:experiments}
\vspace{-0.1in}
\subsection{Experimental Setup}
\label{sec:setup}

We provide a brief overview of our evaluation framework; full details on data sources, return calculations, baseline implementations, hyperparameters, and computational resources are deferred to Appendix~\ref{app:setup}.


\textbf{Dataset and Portfolio.} We evaluate on three distinct 16-stock universes chosen to reflect different financial regimes: \textit{AI Tech} (narrative-driven momentum), \textit{Biotech} (event-driven catalyst jumps), and \textit{Consumer Discretionary} (macro-cycle dependence).
For each trading day from April 2025 to March 2026, the LLM analyst receives the past 1 or 3 days of news articles (filtered by a strict 23:59 UTC cutoff to prevent look-ahead bias), trailing prices, and macro context (SPY, VIX, 10Y Yield).
Crucially, to rigorously prevent data leakage, we evaluate on a 2025--2026 timeline and specifically select GPT-4o-mini and GPT-4.1-mini, whose training cutoffs strictly predate our test data.
Tickers are ranked by the LLM's composite signal to form an \textit{equal-weight, dollar-neutral 5L/5S portfolio}. 
This quantile-based allocation aligns with typical industry practices for evaluating long-short strategies: by concentrating capital on the cross-sectional tails of the prediction distribution, it tests the model's predictive power on the highest-conviction signals while mitigating exposure to the less informative middle, thereby avoiding unnecessary transaction costs on low-conviction predictions. 
We compute \textit{open-to-open (O2O) returns} to realistically model next-day execution, charging 5\,bps transaction costs per trade.

\textbf{Walk-Forward Protocol.} To test out-of-sample generalization, we employ a rolling walk-forward backtest. Each of the three windows consists of:
(1)~a 4-month \textit{training} period where the rubric evolves for $J{=}5$ rounds starting from a shared initial rubric $\mathcal{R}^{(0)}$ of generic financial heuristics (more details in Appendix~\ref{app:seed_rubric}; the attribution agent examines the $K_{\text{attr}}{=}20$ worst days per round);
(2)~a 1-month \textit{validation} period to gate candidate rubrics;
(3)~a 2-month \textit{test} period where the best rubric is frozen and evaluated out-of-sample.

\textbf{Baselines.} We compare \method against: (a) non-LLM statistical baselines (Random L/S, tuned Momentum, tuned Mean Reversion); (b) an ablation using the same LLM but no rubric evolution (\textit{Static rule}); and (c) state-of-the-art LLM-based financial agents, including Lopez-Lira~\citep{lopezlira2023chatgpt}, FinCon~\citep{yu2024fincon}, and FinHEAR~\citep{finhear2025}, all evaluated under identical portfolio constraints. Our primary evaluations use GPT‑4o‑mini and GPT‑4.1‑mini, as mentioned earlier, to prevent data leakage (i.e., the models’ knowledge cutoff predates the training and test dataset timelines). See Appendix~\ref{app:setup} for more details.

\begin{table*}[h]
\centering
\caption{Walk-forward results. Dollar-neutral 5L/5S, 5\,bps cost, O2O returns. Each cell is shaded by its within-column rank (darker $=$ better; rank~1 bold, rank~2 underlined; ranks~8+ left blank). }
\label{tab:summary}
\resizebox{0.89\linewidth}{!}{
\begin{tabular}{@{}l rrrr rrrr rrrr@{}}
\toprule
& \multicolumn{4}{c}{AI Tech} & \multicolumn{4}{c}{Biotech} & \multicolumn{4}{c}{Cons.\ Disc.} \\
\cmidrule(lr){2-5} \cmidrule(lr){6-9} \cmidrule(lr){10-13}
Method & Ret & SR & MaxDD & Cal & Ret & SR & MaxDD & Cal & Ret & SR & MaxDD & Cal \\
\midrule
\multicolumn{13}{@{}l}{\textit{Non-LLM baselines}} \\
Random L/S & $-$8.4 & $-$0.76 & $-$21.1 & $-$0.74 & $-$7.8 & $-$0.84 & $-$17.9 & $-$0.72 & $-$7.7 & $-$0.94 & $-$16.6 & $-$0.68 \\
Mean Reversion & $-$18.4 & $-$1.38 & $-$23.9 & $-$0.77 & $-$17.9 & $-$1.72 & $-$27.7 & $-$0.65 & $-$18.3 & $-$1.82 & $-$24.5 & $-$0.75 \\
Momentum (tuned) & $-$9.5 & $-$0.62 & \rF{$-$16.8} & $-$0.56 & \rD{+12.6} & \rC{+1.19} & \rG{$-$14.8} & \rE{+0.85} & $-$18.0 & $-$2.00 & $-$21.1 & $-$0.85 \\
\midrule
\multicolumn{13}{@{}l}{\textit{LLM-based baselines}} \\
Lopez-Lira\textsubscript{4o}~(\cite{lopezlira2023chatgpt}) & +7.0 & +0.67 & $-$22.7 & +0.31 & $-$14.4 & $-$1.71 & $-$23.4 & $-$0.62 & \rG{$-$3.2} & \rG{$-$0.26} & \rG{$-$10.7} & $-$0.30 \\
Lopez-Lira\textsubscript{4.1}~(\cite{lopezlira2023chatgpt}) & \rD{+15.4} & \rE{+1.24} & $-$19.2 & \rF{+0.80} & $-$10.1 & $-$1.16 & \rE{$-$12.6} & $-$0.80 & $-$7.0 & $-$0.68 & $-$11.8 & $-$0.60 \\
FinCon\textsubscript{4o}~(\cite{yu2024fincon}) & \rE{+14.8} & \rD{+1.41} & \rC{$-$10.0} & \rC{+1.48} & \rA{+18.8} & \rA{+1.75} & \rB{$-$9.8} & \rA{+1.91} & \rB{+14.0} & \rB{+1.62} & \rB{$-$7.3} & \rB{+1.92} \\
FinCon\textsubscript{4.1}~(\cite{yu2024fincon}) & \rG{+8.5} & \rG{+0.80} & \rE{$-$11.0} & \rG{+0.77} & \rE{+10.0} & \rE{+0.99} & \rA{$-$8.0} & \rC{+1.26} & \rC{+8.6} & \rC{+1.06} & \rD{$-$8.4} & \rC{+1.02} \\
FinHEAR\textsubscript{4o}~(\cite{finhear2025}) & \rC{+22.8} & \rB{+1.97} & \rB{$-$9.2} & \rB{+2.48} & \rF{+9.2} & \rF{+0.91} & \rD{$-$12.4} & \rF{+0.74} & \rD{+1.2} & \rD{+0.23} & \rC{$-$7.5} & \rD{+0.16} \\
FinHEAR\textsubscript{4.1}~(\cite{finhear2025}) & \rF{+11.5} & \rF{+1.07} & \rD{$-$10.2} & \rD{+1.12} & $-$5.9 & \rG{$-$0.47} & $-$14.8 & \rG{$-$0.40} & \rF{$-$2.2} & \rF{$-$0.16} & \rE{$-$8.9} & \rF{$-$0.25} \\
\midrule
\multicolumn{13}{@{}l}{\textit{\method (ours)}} \\
\method\textsubscript{4o} & \rA{+33.2} & \rA{+2.45} & \rA{$-$7.3} & \rA{+4.57} & \rC{+12.7} & \rB{+1.26} & \rC{$-$11.5} & \rD{+1.11} & \rA{+16.8} & \rA{+1.77} & \rA{$-$7.2} & \rA{+2.32} \\
\method\textsubscript{4.1} & \rB{+22.9} & \rC{+1.44} & $-$27.3 & \rE{+0.84} & \rB{+17.8} & \rD{+1.08} & \rF{$-$13.5} & \rB{+1.32} & \rE{$-$0.5} & \rE{+0.10} & \rF{$-$9.8} & \rE{$-$0.05} \\
\bottomrule
\end{tabular}
}
\vspace{-0.1in}
\end{table*}

\subsection{Main Results}
\label{sec:main_results}

\cref{tab:summary} compares the main methods across three sectors, each evaluated on four metrics: total return, Sharpe ratio, maximum drawdown, and Calmar ratio.
For each LLM backbone, we select the best-performing news window per sector using only the validation month of each walk-forward split, then freeze that choice for the test months.
All test periods are strictly out-of-sample under the walk-forward protocol.

To keep \textit{no signal} honest in the presence of 5\,bps costs, we compare every method against a 1000-trial Random L/S null (Appendix~\ref{app:setup}; 3-sector mean return $-$8.0\%, Sharpe $-$0.85). The null is negative because random daily rebalancing incurs high turnover (including directional flips counted as two trades each), so the cost compounds to ${\sim}8\%$ of drag over the 122-day test; this is the correct reference, not zero.
Because the study contains only three non-overlapping walk-forward windows, we interpret the results as directional out-of-sample evidence rather than a formal significance claim about long-term excess returns.

On 3-sector average, the evolved rubrics clearly separate from this null. Rubric evolution lifts SHARP\textsubscript{4o} from its Static initialization ($+0.6\%$ / $+0.09$\,SR, barely above the Random null) to $+20.9\%$ / $+1.83$\,SR, and SHARP\textsubscript{4.1} from $-4.4\%$ / $-0.36$ to $+13.4\%$ / $+0.87$. The remaining baselines exceed the Random null but remain economically negligible: tuned Momentum averages $-5.0\%$, Lopez-Lira $-0.6\%$ to $-3.5\%$, and FinHEAR\textsubscript{4.1} $+1.1\%$---none meaningful after costs, and even Momentum's $+1.19$\,SR in Biotech does not survive aggregation across regimes. The closest contender is FinCon\textsubscript{4o}, whose conceptual-verbal reinforcement reaches $+15.9\%$ on 3-sector average and outperforms SHARP\textsubscript{4o} on Biotech ($+18.8\%$ vs $+12.7\%$). SHARP\textsubscript{4o} nonetheless leads the 3-sector average by roughly $5$\,pp. FinCon's free-form reinforcement is comparatively better matched to the event-driven structure of Biotech. Even the one SHARP cell that comes in slightly negative, SHARP\textsubscript{4.1} on Consumer Disc.\ at $-0.5\%$, is a large lift over its Static initialization ($-12.9\%$, Appendix~\ref{app:static}) and improves on the Random null ($-7.7\%$), suggesting that evolution provides a consistent directional correction even when the backbone is weaker in a given sector.

\begin{wrapfigure}{r}{0.65\textwidth} 
  \vspace{-15pt} %
  \centering

  \includegraphics[width=\linewidth]{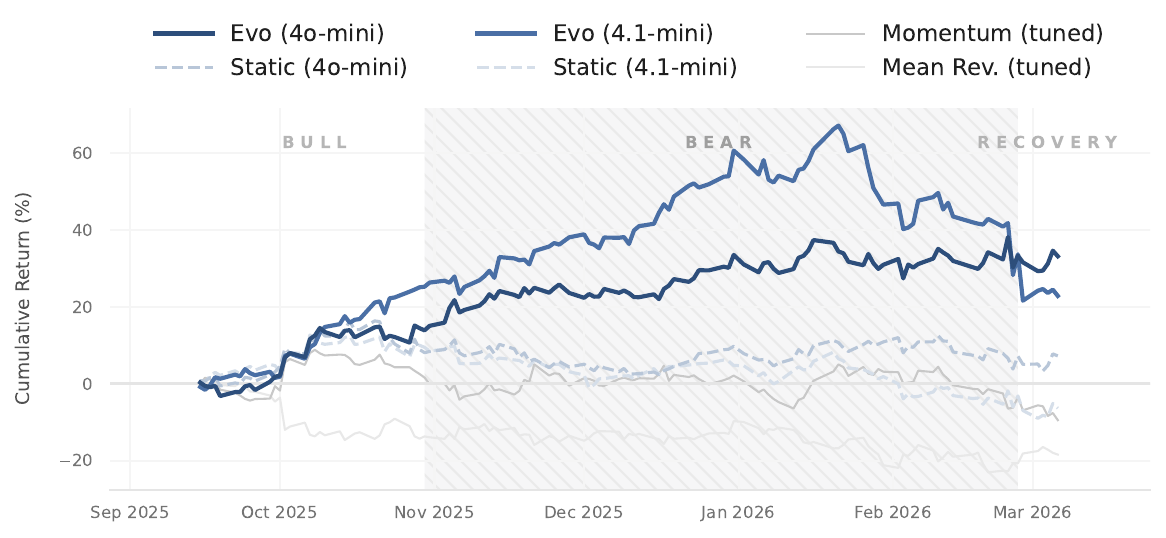}
  \caption{Cumulative OOS returns, AI Tech. Evolved rubrics (solid) steadily separate from their static counterparts (dashed); momentum and mean reversion remain flat or negative.}
  \label{fig:cumret}
  \vspace{-15pt} 
\end{wrapfigure}

\cref{fig:cumret} illustrates the cumulative return trajectories for the AI Tech sector.
The evolved-rubric variants (solid lines) steadily diverge from their static-rule counterparts (dashed lines) over the test period, confirming that the gains generalize across the evaluated periods.
While GPT-4.1-mini + Evo exhibits a steeper maximum drawdown ($-$27.3\%) than GPT-4o-mini ($-$7.3\%), this drawdown primarily occurs in the later half of the evaluation period, reflecting natural signal decay as the market environment drifts further from the training window.
\vspace{-0.1in}
\subsection{What Does the Rubric Learn?}
\label{sec:analysis}
\vspace{-0.1in}
By inspecting the rules that emerge across evolution rounds, we find that the system dynamically overrides generic priors (e.g., flipping mean-reversion to momentum in AI Tech) to match underlying market micro-structures, adapting its analysis framework to sector-specific dynamics.
\cref{fig:diffs} shows representative before/after diffs for AI Tech (see Appendix~\ref{app:diffs} for Biotech and Consumer Discretionary).

\begin{wrapfigure}{r}{0.55\textwidth}
\vspace{-12pt}
\begin{Verbatim}[commandchars=\\\{\},fontsize=\tiny]
\textcolor{diffgray}{@@ temporal_priced_in @@}
\textcolor{diffredtext}{- IF: moved >3% in news direction THEN: Reduce signal by 70%}
\textcolor{diffgreentext}{+ IF: moved >5% AND news positive THEN: Increase expected return}
\textcolor{diffgray}{@@ macro_high_vix @@}
\textcolor{diffredtext}{- IF: VIX above 25          THEN: Cut bullish returns by 30%}
\textcolor{diffgreentext}{+ IF: VIX above 25, falling THEN: Maintain bullish returns}
\textcolor{diffgray}{@@ news_generic_market @@}
\textcolor{diffredtext}{- IF: generic market news THEN: Reduce weight by 60%}
\textcolor{diffgreentext}{+ IF: generic market news THEN: Reduce weight by 80%}
\end{Verbatim}
\vspace{-4pt}
\caption{AI Tech diffs. Evolution discovers momentum, refines VIX interactions, and prunes filters. Full diffs in Appendix~\ref{app:diffs}.}
\label{fig:diffs}
\vspace{-8pt}
\end{wrapfigure}

Three interpretable patterns emerge from these edits:
(1)~\textit{Discovering return dynamics}: the \texttt{temporal\_priced\_in} initial rule (a mean-reversion prior) is flipped to a momentum signal in AI Tech to capture the 2025 AI narrative rally, narrowed to a conditional exception in Consumer Discretionary (preserving discount in most cases but suppressing it on contrarian rebounds), and specialized strictly for FDA events in Biotech.
(2)~\textit{Creating catalyst rules}: evolution introduces directional asymmetry for FDA approvals/rejections, adds a symmetric high-VIX rule in Consumer Discretionary (damping bullish \emph{and} amplifying bearish signals during fear spikes), and refines VIX interactions in AI Tech to buy the dip when volatility is elevated but falling.
(3)~\textit{Tightening noise filters}: across all sectors, the system lowers confidence caps for stocks with scarce news and heavily penalizes generic market commentary.
Crucially, every rule change is a human-readable condition/action edit traceable to specific historical error days. A practitioner can inspect the evolved rubric, verify that each rule aligns with financial intuition, and manually override specific behaviors, a level of transparency that free-form prompt optimization does not offer.
\vspace{-0.1in}
\subsection{Ablation Studies}
\label{sec:ablation}
\begin{wraptable}{r}{0.42\textwidth}
\vspace{-12pt}
\centering
\caption{Ablation on AI Tech. Each row modifies one component vs.\ full system.}
\label{tab:ablation}
\footnotesize
\setlength{\tabcolsep}{1.5pt}
\begin{tabular}{@{}l rr rr@{}}
\toprule
& \multicolumn{2}{c}{GPT-4o-mini} & \multicolumn{2}{c}{GPT-4.1-mini} \\
\cmidrule(lr){2-3} \cmidrule(lr){4-5}
Variant & Ret\,(\%) & SR & Ret\,(\%) & SR \\
\midrule
Full system ($J{=}5$) & \textbf{+33.2} & \textbf{+2.45} & \textbf{+22.9} & +1.44 \\
\midrule
A1: No attribution & +7.1 & +0.73 & +8.5 & +0.69 \\
A2: Free reflection & $-$12.1 & $-$0.84 & +6.0 & +0.60 \\
\midrule
Static (no evolution) & +7.3 & +0.75 & $-$6.4 & $-$0.42 \\
\bottomrule
\end{tabular}
\vspace{-8pt}
\end{wraptable}

We isolate the contribution of \method's components via ablations on the AI Tech sector (\cref{tab:ablation}).
The results confirm that the \textit{attribution agent (A1)} is the most critical component: without error-day analysis, mutations become untargeted, and performance drops to near-static levels.
Replacing the structured rule-editing step with \textit{free-form reflection (A2)}, which serves as our \emph{vanilla prompt optimization} baseline, causes an even sharper degradation.
The A2 variant is initialized with the identical rubric $\mathcal{R}^{(0)}$ and keeps every other component identical (the same LLM backbone, attribution signals, evolution budget, and validation gate), changing only the \emph{mutation operator}: candidates are produced by rewriting the analyst prompt in free form instead of emitting atomic rule edits.
Under this change, GPT-4o-mini's Sharpe drops from $+2.45$ to $-0.84$ and total return drops from $+33.2\%$ to $-12.1\%$, notably \emph{below} the Static baseline that performs no adaptation at all.
This empirical collapse directly illustrates the \emph{credit assignment problem} discussed in \cref{sec:intro}. When edits rely on unbounded text rewrites rather than structured fault isolation (as is standard in vanilla prompt optimization), each optimization round can silently overwrite previously correct behavior (a phenomenon known as policy degeneration) in ways that the validation gate cannot fully catch.
Qualitatively, even when fair-initialized with the same rubric content, the A2 prompt gradually shifts from modular \texttt{IF/THEN} rules into flowing prose whose thresholds drift without any traceable link to specific error days (Appendix~\ref{app:reflect_example}).
\method's evolved rubric, by contrast, remains a bounded condition/action list whose every change is a single-rule diff attached to the attribution pattern that motivated it (\cref{fig:diffs}).

\noindent\textbf{Open-source LLM backbones (A3).} To test whether \method depends on proprietary APIs, we evaluate two open-source models, Qwen2.5-72B-Instruct~\cite{qwen2025qwen25technicalreport} and Llama-3.3-70B-Instruct~\cite{grattafiori2024llama}, under the same walk-forward protocol.
\cref{tab:opensource} reports the three-sector average return and Sharpe ratio alongside non-LLM baselines and the static-rule ablation.
\begin{wraptable}{r}{0.3\textwidth}
\vspace{-9pt}
\centering
\caption{Open-source backbones: three-sector average.}
\vspace{-5pt}
\label{tab:opensource}
\footnotesize
\setlength{\tabcolsep}{2pt}
\begin{tabular}{@{}l rr@{}}
\toprule
Method & Ret\,(\%) & SR \\
\midrule
Random L/S & $-$8.0 & $-$0.85 \\
MeanRev (tuned) & $-$18.2 & $-$1.64 \\
Momentum (tuned) & $-$5.0 & $-$0.48 \\
\midrule
Qwen-72B + Static & $-$0.6 & $-$0.10 \\
Qwen-72B + Evo & +9.2 & +0.88 \\
\midrule
Llama-70B + Static & $-$5.3 & $-$0.69 \\
Llama-70B + Evo & +7.6 & +0.72 \\
\bottomrule
\end{tabular}
\vspace{-8pt}
\end{wraptable}
The pattern mirrors the proprietary backbones: with the initial rubric alone, both open-source models deliver marginal returns ($-$0.6\% / $-$5.3\%) that, while above the Random null ($-$8.0\%), remain economically negligible.
Rubric evolution closes this gap cleanly, lifting Qwen-72B by $+9.8$\,pp (SR from $-$0.10 to +0.88) and Llama-70B by $+12.9$\,pp (SR from $-$0.69 to +0.72), so that both evolved variants turn clearly profitable and exceed every non-LLM baseline.
The Static-to-Evo lift is comparable in magnitude to the proprietary backbones, so the contribution of rubric evolution is attributable to the mechanism itself rather than to any specific LLM family, and \method transfers to weights that can be hosted on-premise.

\vspace{-0.1in}
\section{Discussion: From Backtest to Production}
\label{sec:discussion}
\vspace{-0.12in}
The results in \cref{sec:experiments} demonstrate that structured rubric evolution can produce profitable, interpretable trading signals in a controlled walk-forward backtest.
We view these findings as a \emph{feasibility demonstration} rather than a production-ready system.
The current implementation uses a generic initial rubric, unconstrained rule mutations, and a single public news source.
Future work could integrate richer data streams available to institutional practitioners, including premium analyst reports, proprietary alternative data, and limit order book (LOB) micro-structure.
While these high-dimensional signals would increase the complexity of the evolution process, they also represent a substantial ceiling for further performance gains, as SHARP's current outperformance is achieved using only a sparse set of public news features.
A realistic assessment of the gap between the backtested performance reported here and realized live returns is therefore essential.

\noindent\textbf{Gap to production.}
Several factors separate backtested performance from live trading returns.
(i)~\emph{Execution slippage.} Our backtest assumes rebalancing at official open-to-open prices, which is optimistic relative to a live implementation. In practice, a production system would usually need to execute over a short post-open window rather than transact the full portfolio at the auction print, so realized fills would drift away from the benchmark open, especially in smaller names and on high-volatility news days.
(ii)~\emph{Market impact.} Our 5\,bps transaction cost covers commissions but omits the price impact of the portfolio's own trades. While negligible for highly liquid AI Tech names, executing trades in less liquid sectors, such as Biotech or Consumer Discretionary, incurs a material market impact that inherently bounds the strategy's capacity \citep{almgren2005direct}.
(iii)~\emph{Signal decay.} Prior work shows that text-based news signals are short-lived, often incorporated into prices within a window of one to four days \citep{tetlock2007giving, heston2016news, ke2019predicting}. Our daily rebalancing therefore, relies on the overnight signal retaining value through the open, which is plausible for major earnings or regulatory events but less so for routine commentary.
(iv)~\emph{Regime dependence.} Our seven-month out-of-sample window spans a rally and a modest correction, but is too short to assess robustness to structural shifts such as prolonged bear markets or abrupt monetary policy changes. The sector-specific adaptations discovered by evolution may themselves be regime-dependent, necessitating periodic re-evolution.
(v)~\emph{Operational risk.} Deploying LLM signals introduces dependencies on API availability, news feed latency, and model versioning. A delayed inference before market open leaves the portfolio without a signal, and each quarterly re-evaluation cycle requires human review before acceptance.

\noindent\textbf{Human oversight and the rubric as an audit interface.}
In practice, no quantitative trading strategy operates without continuous human oversight, regardless of whether it is built on statistical factors, deep learning, or LLMs.
Risk managers, portfolio managers, and compliance officers must be able to evaluate model behavior, diagnose failures, and intervene when market conditions change.
Traditional ML-based alpha models make this difficult: a gradient-boosted tree over 200 engineered features or a transformer trained end-to-end on order-book data offers limited surface area for a human to understand why a particular position was taken, let alone to selectively override one component of the model's reasoning without retraining.
The rubric representation in \method provides a natural interface for this oversight process.
Each rule is a human-readable condition/action pair grounded in financial concepts, such as the interaction between VIX levels and bullish signals, and the evolution history records which error days motivated each rule change.
A practitioner can review the evolved rubric before deployment, agree with some rules, override others based on proprietary views or risk limits, and do so without touching the underlying LLM or rerunning the optimization.
This \emph{selective editability} distinguishes rubric-based systems from both black-box ML and free-form prompt optimization, where any manual intervention risks unpredictable downstream effects.

\noindent\textbf{On the role of the LLM.}
It is worth being explicit about what the LLM does and does not contribute in our framework.
We do not claim that LLMs perform genuine financial reasoning; whether current language models truly understand market dynamics or are performing sophisticated pattern matching on their training corpora remains an open question.
The rubric framework is designed so that this distinction matters less than it otherwise would: the financial judgment is encoded in the rubric and refined through market feedback, while the LLM's role is closer to structured text comprehension, mapping unstructured news into the rubric's condition space.
When the LLM misclassifies a news article or applies a rubric condition incorrectly, the error surfaces as a traceable mismatch between a specific rule and its outcome, rather than as an opaque failure buried in the model's parameters.
This separation does not eliminate the dependence on LLM quality, but it does refocus the LLM's contribution to a narrower, more verifiable task than end-to-end return prediction. Notably, the evolutionary loop ensures that these rules are not static biases from the prompt but are empirically validated and tuned against realized P\&L.

\section{Conclusion}
\label{sec:conclusion}
We introduced \method, an LLM-based trading framework that addresses the credit assignment problem in self-improving agents by replacing unconstrained prompt mutation with structured rubric evolution. 
Empirically, \method turns generic initial rubrics into systematic walk-forward outperformance across three equity sectors, lifting the performance while maintaining structural transparency and auditability. 
The primary limitations of our study include the omission of execution slippage, market impact at scale, and long-term regime drift (\cref{sec:discussion}), which we leave for future work alongside expanding the rubric schema. 
Ultimately, \method represents a shift in paradigm for quantitative AI: the LLM is confined to mapping unstructured text into a condition space, while the financial judgment itself is held securely in the rubric and iteratively refined from market feedback, allowing the agent to correct human-provided priors when they mismatch the current regime.

\section*{Disclaimer}
This material is made available for informational purposes only. Morgan Stanley and its affiliates (“Morgan Stanley”) makes no representation and warranty whatsoever and disclaims all liability, for the completeness, accuracy or reliability of the information contained herein. This material is not investment research or investment advice, or a recommendation, offer or solicitation for the purchase or sale of any security, financial instrument, financial product or service, or to be used in any way for evaluating the merits of participating in any transaction or trading strategy. This material was not prepared by the Morgan Stanley Research Department. The information contained in this material does not constitute advice or Morgan Stanley research. Morgan Stanley is not acting as your advisor (municipal, financial, or otherwise) and is not acting in a fiduciary capacity. Unless otherwise indicated, any views expressed or information contained in the materials are those of the respective author(s) alone and may differ from others within Morgan Stanley, including those of Morgan Stanley's Research Department or sales and trading groups. Morgan Stanley did not conduct an independent review of the materials and may not utilize the information, analysis and/or strategies provided by the author(s). Past performance is not indicative of future returns.

\bibliographystyle{unsrt}
\bibliography{references}

\appendix

\section*{Broader Impacts}

The strongest case for \method is not full automation, but a shorter loop between \emph{observing} a failure mode and \emph{editing} the policy that caused it.
Because adaptation is expressed as explicit rule edits, changes can be inspected and selectively accepted before capital is put at risk.
Compared with conventional black-box pipelines, this makes regime adaptation more governable: diagnosis, revision, and review happen in the same representation rather than across disconnected model and prompt layers.

\section{Detailed Experimental Setup}
\label{app:setup}

This section provides the full technical details underlying \cref{sec:setup}.

\noindent\textbf{Stock universes.}
We construct three sector-specific universes of 16 stocks each, chosen to test \method under qualitatively different signal regimes:
\begin{itemize}[nosep,leftmargin=*]
\item \textbf{AI \& Tech} (16): NVDA, MSFT, GOOGL, META, AMZN, AAPL, AMD, AVGO, TSM, MRVL, ARM, CRM, PLTR, SNOW, AI, TSLA. Spans mega-cap AI leaders, semiconductor infrastructure, and AI software. This sector is \emph{narrative-driven}: prices move with macro sentiment, earnings guidance, and AI hype cycles.
\item \textbf{Biotech} (16): LLY, JNJ, ABBV, MRK, PFE, AMGN, GILD, BMY, VRTX, REGN, MRNA, BIIB, ISRG, DXCM, ILMN, ALNY. Spans large-cap pharma and mid-cap biotech. This sector is \emph{event-driven}: prices jump on binary catalysts such as FDA approvals, clinical trial readouts, and drug pipeline updates.
\item \textbf{Consumer Discretionary} (16): AMZN, TJX, ROST, TSLA, HD, LOW, DHI, LEN, MCD, SBUX, CMG, NKE, BKNG, ORLY, AZO, GRMN. Covers e-commerce, automotive, home improvement, restaurants, apparel, and travel. This sector is \emph{cycle-driven}: prices track consumer spending, housing data, and macro conditions.
\end{itemize}

\noindent\textbf{Data sources.}
\textit{Prices.} Daily OHLCV data from Yahoo Finance for all 16 tickers per sector, plus three macro indicators: SPY (S\&P~500 ETF), VIX (CBOE Volatility Index), and the 10-Year Treasury Yield. The price dataset spans Apr~2024 to Mar~2026: the first ${\sim}$12 months (Apr~2024 to Apr~2025) serve solely as lookback context for 52-week high/low and 20-day momentum features, while the trading evaluation period (training through testing) runs from Apr~2025 onward.

\textit{News.} Historical news articles sourced from Finnhub and cached on disk for reproducibility. For each trading day~$T$, we fetch articles for each ticker from the preceding $d$ calendar days, where $d= 3$ is the \emph{news window}. Every article is filtered by its \texttt{published} timestamp: we discard any article timestamped after $T$\,23:59\,UTC. Since the US market opens at 09:30\,ET (13:30\,UTC), this creates a \textbf{13.5-hour safety gap} between the news cutoff and the next morning's execution, preventing all look-ahead bias from after-hours or overnight articles.

\textit{Price context.} For each ticker on each day~$T$, the LLM receives: last close price, 1-day/5-day/20-day returns, 52-week high and low (computed from the trailing 252 trading days ending at~$T$, strictly backward-looking), and percent distance within this range. Macro context includes the same momentum features for SPY, VIX, and the 10-Year yield. All context is available at market close on day~$T$; no forward-looking information is used.

\noindent\textbf{Return calculation.}
We use \textbf{open-to-open (O2O) returns} to model realistic execution:
\begin{enumerate}[nosep,leftmargin=*]
\item \textit{Decision time}: day $T$ close. The LLM has access to all prices up to $T$ close and all news up to $T$\,23:59\,UTC.
\item \textit{Execution}: a market-on-open (MOO) order fires at day $T{+}1$ open (09:30\,ET).
\item \textit{Return}: $r_i = \textrm{Open}_i^{T+2} / \textrm{Open}_i^{T+1} - 1$ for each held position~$i$.
\end{enumerate}
This avoids the common pitfall of using close-to-close returns with same-day signals, which implicitly assumes execution at an already-known close price.

\noindent\textbf{Portfolio construction.}
For each trading day~$T$, the LLM produces a composite signal for each ticker: $\sigma_i = \hat{r}_i \cdot c_i$, where $\hat{r}_i$ is the LLM's expected return estimate and $c_i \in [0, 1]$ is its confidence.
Tickers are ranked by $\sigma_i$; we go \textbf{long the top~5} and \textbf{short the bottom~5}, forming an \textbf{equal-weight, dollar-neutral} 5L/5S portfolio rebalanced daily.
The 5L/5S constraint is always binding: because every stock receives a score $\sigma_i$, the top and bottom five are always well-defined regardless of confidence levels, so no fill or reweighting logic is needed.
Each leg allocates $\tfrac{1}{5}$ of its notional to every constituent; the daily portfolio return is $r^{\text{long}} + r^{\text{short}}$, where $r^{\text{long}} = \tfrac{1}{5}\sum_{i \in L} r_i$ and $r^{\text{short}} = -\tfrac{1}{5}\sum_{j \in S} r_j$, using open-to-open (O2O) returns.
Transaction costs of 5 basis points are charged for each new position entering the portfolio (i.e., each time a stock enters the long or short leg that was not there the previous day).

\noindent\textbf{Walk-forward protocol details.}
We employ a rolling walk-forward backtesting protocol with the following window structure:
\begin{itemize}[nosep,leftmargin=*]
\item \textbf{Train} (4 months): the rubric evolves for $J{=}5$ rounds. In each round, the attribution agent identifies error days from the training-period backtest, the evolution agent proposes $\leq$3 rule mutations, and the candidate rubric is evaluated on validation data.
\item \textbf{Validation} (1 month): each candidate rubric is backtested on validation data. A candidate is \emph{accepted} if its validation excess return satisfies $r_{\mathrm{val}} > r_{\mathrm{val}}^{*} - 0.005$, where $r_{\mathrm{val}}^{*}$ is the best seen so far. This 0.5\,pp tolerance avoids rejecting sideways improvements. The rubric with the highest $r_{\mathrm{val}}$ across all rounds is selected.
\item \textbf{Test} (2 months): the best rubric is frozen and evaluated on strictly out-of-sample data. No rubric changes, no parameter tuning, and no information from the test period leaks into any earlier stage.
\item \textbf{Step}: the window advances by 2 months, so consecutive test periods tile without overlap.
\end{itemize}
With price data spanning Apr~2024 to Mar~2026 and the trading evaluation period beginning Apr~2025, this yields \textbf{3 walk-forward windows} per sector, with test periods covering approximately Sep~2025 to Mar~2026 (${\sim}$122 trading days total). Within each window, rubric evolution is \emph{independent}: the same initial rubric $\mathcal{R}^{(0)}$ is used as the starting point, and all 5 rounds of evolution occur on that window's train+val data only.

\noindent\textbf{Baselines.}
We compare against four baselines that span three levels of sophistication:
\begin{enumerate}[nosep,leftmargin=*]
\item \textbf{Random L/S}: 1{,}000 Monte Carlo trials. Each trial randomly selects 5 long and 5 short positions from the same 16-stock universe, rebalanced daily with the same 5\,bps cost. Transaction costs account for directional flips (e.g., a stock moving from the long to the short leg counts as two trades: closing the old position and opening the new one). Although the position construction is dollar-neutral, the expected return is \emph{not} zero: high random turnover and frequent directional flips cause the 5\,bps per-trade cost to compound to roughly $-$8\% over the 122-day test window (three-sector mean $-$8.0\%, mean Sharpe $-$0.85). This null is therefore the correct reference for ``no real signal'' under our execution assumptions, rather than a naive zero.
\item \textbf{Momentum (tuned)}: rank tickers by $k$-day return; long top-5, short bottom-5. The lookback $k$ is selected from $\{1, 2, 3, 5, 10, 20\}$ days by maximizing Sharpe on the combined train+validation data \emph{before} evaluating on test. This respects the train/test boundary.
\item \textbf{Mean Reversion (tuned)}: rank tickers by \emph{negative} $k$-day return (buy losers, short winners). Same lookback selection procedure as momentum.
\item \textbf{Static rule (no evolution)}: the same LLM backbone receives the same news, price context, and macro data, and uses the initial rubric $\mathcal{R}^{(0)}$ with no evolution applied. The LLM produces $(\hat{r}_i, c_i)$ signals guided only by the initial rules, without any sector-specific adaptation from P\&L feedback. This is the key ablation: Static and Evo share the same initialization and differ \emph{only} in whether evolution is run.
\end{enumerate}

\noindent\textbf{News window.}
For each trading day~$T$, the LLM analyzes news from the preceding $d$ calendar days. In the main experiments, $d = 3$; when multiple windows are compared, the choice is made using only the validation month of each walk-forward split and then frozen for test.

\noindent\textbf{Evaluation metrics.}
We report four metrics: \textbf{total return} (\%), \textbf{Sharpe ratio} (annualized: $\mu / \sigma \times \sqrt{252}$), \textbf{maximum drawdown} (peak-to-trough decline, \%), and \textbf{Calmar ratio} ($\text{Total Return} / |\text{MaxDD}|$).

\noindent\textbf{Implementation details.}
For rubric evolution (\cref{sec:evolution}): $J{=}5$ rounds per walk-forward window, maximum rubric size $M_{\max} = 18$ rules (hard cap enforced programmatically), at most $\Delta_{\max} = 3$ mutations per round, validation tolerance $\epsilon = 50$\,bps.
The attribution agent examines the $K_{\text{attr}}{=}20$ worst-performing portfolio days and filters for error patterns appearing on $\geq 3$ days.
To encourage compactness, the evolution agent's prompt instructs it to remove an existing rule for each addition once the rubric reaches $\geq 12$ rules; the hard upper bound of $M_{\max} = 18$ is enforced in code and is rarely approached in practice.
The initial rubric (round~0) contains a small set of generic rules and is shared across all sectors; it is not tuned on any test-period data.

\noindent\textbf{Computational resources and runtime.}
Proprietary backbones (GPT-4o-mini, GPT-4.1-mini) are accessed via the OpenAI API and require no local GPUs.
Open-source backbones (Qwen2.5-72B-Instruct, Llama-3.3-70B-Instruct) are served locally with vLLM (\texttt{tensor\_parallel\_size}~$=$~4, \texttt{bfloat16}, \texttt{max\_model\_len}~$=$~32\,768) on a single node of $8{\times}$NVIDIA~H100~80GB; we allocate 4~GPUs per model, so the two open-source backbones can be served concurrently.
A complete per-sector run, covering all three walk-forward windows, $J{=}5$ evolution rounds per window, and both the Static and Evo evaluation paths (so that all attribution backtests, evolution proposals, and validation backtests are included), has the following measured wall-clock cost (averages over the runs logged for \cref{tab:summary}):
(i)~$\sim$3\,h~30\,min for GPT-4o-mini / GPT-4.1-mini on the OpenAI API;
(ii)~$\sim$5\,h~40\,min for Qwen-72B on $4{\times}$H100; and
(iii)~$\sim$6\,h~55\,min for Llama-70B on $4{\times}$H100.

\section{Evolution Algorithm}
\label{app:algorithm}

\cref{alg:sharp} gives the full training loop for rubric evolution: backtest, attribute errors, propose candidate rule edits, and validate before accepting.

\begin{algorithm}[htbp]
\caption{\method Rubric Evolution (Training Phase)}
\label{alg:sharp}
\begin{algorithmic}[1]
\REQUIRE $\mathcal{D}_{\text{train}}$, $\mathcal{D}_{\text{val}}$, initial rubric $\mathcal{R}^{(0)}$, rounds $J$, tolerance $\epsilon$
\STATE $\mathcal{R}^* \leftarrow \mathcal{R}^{(0)}$,\; $e^* \leftarrow e\bigl(\textsc{Backtest}(f_\theta(\cdot \mid \mathcal{R}^{(0)}),\; \mathcal{D}_{\text{val}})\bigr)$
\FOR{$j = 1$ \TO $J$}
    \STATE $\mathbf{y} \leftarrow \textsc{Backtest}\bigl(f_\theta(\cdot \mid \mathcal{R}^{(j-1)}),\; \mathcal{D}_{\text{train}}\bigr)$
    \STATE $\mathcal{E} \leftarrow \textsc{Attribute}\bigl(\text{worst days of } \mathbf{y}\bigr)$
    \STATE $\mathcal{P} \leftarrow \textsc{Evolve}\bigl(\mathcal{R}^{(j-1)},\; \mathcal{E}\bigr)$ \COMMENT{candidate rule edits}
    \STATE $\tilde{\mathcal{R}}^{(j)} \leftarrow \textsc{Apply}\bigl(\mathcal{R}^{(j-1)},\; \mathcal{P}\bigr)$
    \STATE $e_j \leftarrow e\bigl(\textsc{Backtest}(f_\theta(\cdot \mid \tilde{\mathcal{R}}^{(j)}),\; \mathcal{D}_{\text{val}})\bigr)$ \COMMENT{validation score}
    \IF{$e_j \geq e^* - \epsilon$}
        \STATE $\mathcal{R}^{(j)} \leftarrow \tilde{\mathcal{R}}^{(j)}$
        \IF{$e_j > e^*$}
            \STATE $\mathcal{R}^* \leftarrow \tilde{\mathcal{R}}^{(j)}$,\; $e^* \leftarrow e_j$
        \ENDIF
    \ELSE
        \STATE $\mathcal{R}^{(j)} \leftarrow \mathcal{R}^{(j-1)}$
    \ENDIF
\ENDFOR
\RETURN $\mathcal{R}^*$
\end{algorithmic}
\end{algorithm}

\section{Initial Rubric Template}
\label{app:seed_rubric}

The initial rubric $\mathcal{R}^{(0)}$ is a small generic set of rules used to start the evolution process.
No rule is tuned on test-period data; all conditions and thresholds are adapted from standard financial commentary available prior to our evaluation window.
The following six rules are shared across all sector universes.

\begin{Verbatim}[fontsize=\scriptsize,frame=single,commandchars=\\\{\}]
\textbf{[temporal_discount] temporal_priced_in}
  Discount news already reflected in price.
  IF: Stock moved >3% in the same direction as the news sentiment
      over the past 5 trading days
  THEN: Reduce the signal strength from this news by 70%

\textbf{[temporal_discount] temporal_earnings_season}
  Downweight non-earnings news near earnings.
  IF: Within 5 trading days before or after a scheduled earnings report
  THEN: Reduce weight of non-earnings-related news by 50%

\textbf{[news_weighting] news_analyst_rating}
  Moderate weight for analyst rating changes.
  IF: News is an analyst upgrade/downgrade or price target change
  THEN: Base signal weight x1.3 if from top-tier bank, x1.0 otherwise

\textbf{[news_weighting] news_generic_market}
  Downweight generic market commentary.
  IF: News is broad market commentary not specific to the company
  THEN: Reduce signal weight by 60%

\textbf{[macro_interaction] macro_high_vix}
  Dampen bullish signals in high-fear environment.
  IF: VIX is above 25
  THEN: Reduce all bullish expected_return estimates by 30%

\textbf{[news_weighting] news_count_low}
  Lower confidence when news is scarce.
  IF: Fewer than 3 news articles available for the ticker
  THEN: Cap confidence at 0.4
\end{Verbatim}

Each sector additionally includes 1--4 sector-specific rules (e.g., FDA event weighting for biotech, housing cycle rules for consumer discretionary); these are listed in full in the released code.

\section{Static Rule Baseline (Full Results)}
\label{app:static}

\cref{tab:static_full} reports the full static-rule baseline: same LLM backbone and inputs as \method, but no rubric evolution.
These rows are omitted from the main table for space.
Evo vs.\ Static is the key ablation because the two settings differ \emph{only} in whether the rubric has been optimized.

\begin{table}[htbp]
\centering
\caption{Static rule baseline (no evolution): walk-forward results across all models and sectors.}
\label{tab:static_full}
\footnotesize
\setlength{\tabcolsep}{2.5pt}
\begin{tabular}{@{}l rrrr rrrr rrrr@{}}
\toprule
& \multicolumn{4}{c}{AI Tech} & \multicolumn{4}{c}{Biotech} & \multicolumn{4}{c}{Cons.\ Disc.} \\
\cmidrule(lr){2-5} \cmidrule(lr){6-9} \cmidrule(lr){10-13}
Method & Ret & SR & MaxDD & Cal & Ret & SR & MaxDD & Cal & Ret & SR & MaxDD & Cal \\
\midrule
GPT-4o-mini & +7.3 & +0.75 & $-$12.0 & +0.61 & $-$8.7 & $-$0.91 & $-$17.4 & $-$0.50 & +3.1 & +0.44 & $-$8.4 & +0.37 \\
GPT-4.1-mini & $-$6.4 & $-$0.42 & $-$19.2 & $-$0.33 & +6.2 & +0.80 & $-$12.7 & +0.48 & $-$12.9 & $-$1.47 & $-$16.8 & $-$0.77 \\
Qwen2.5-72B & +10.0 & +0.85 & $-$9.8 & +1.02 & $-$7.1 & $-$0.72 & $-$11.4 & $-$0.62 & $-$4.7 & $-$0.42 & $-$14.4 & $-$0.33 \\
Llama-3.3-70B & +12.1 & +1.06 & $-$15.4 & +0.78 & $-$19.3 & $-$2.20 & $-$20.8 & $-$0.93 & $-$8.6 & $-$0.94 & $-$13.5 & $-$0.64 \\
\bottomrule
\end{tabular}
\begin{flushleft}
\footnotesize Returns and MaxDD in \%. Same portfolio construction as \cref{tab:summary}.
\end{flushleft}
\end{table}

\section{Sector-Specific Rubric Diffs}
\label{app:diffs}

\cref{fig:diffs_app} provides representative before/after diffs for the Biotech and Consumer Discretionary sectors, complementing the AI Tech example in the main text (\cref{fig:diffs}).
The same three patterns from the main text reappear here in sector-specific form.

\noindent\textbf{(1) Discovering distinct return dynamics.}
In \textbf{AI Tech}, the \texttt{temporal\_priced\_in} rule flips from a mean-reversion prior into a momentum rule, matching the narrative-driven rally regime of 2025.
In \textbf{Consumer Disc.}, the same rule is narrowed into an exception: the blanket $70\%$ discount is suppressed when a stock moved opposite to positive news, preserving contrarian rebound setups.
In \textbf{Biotech}, the rule is specialized to FDA events with a higher $7\%$ threshold and a milder $50\%$ discount, reflecting the fact that large post-FDA moves are often fundamental rather than noise.

\noindent\textbf{(2) Sector-specific catalyst rules.}
Beyond adapting shared initial rules, evolution also creates new rules tailored to each sector's signal structure.
In \textbf{Biotech}, the FDA weighting rule is strengthened from $2.0{\times}$ to $2.5{\times}$ and made directional: approvals are bullish, rejections bearish.
In \textbf{Consumer Disc.}, evolution adds a symmetric high-VIX rule: when VIX exceeds 25, bullish predictions are damped and bearish predictions amplified by $30\%$.

\begin{figure*}[h]
\textbf{(a) Biotech} \hfill {\scriptsize FDA amplified with asymmetry; priced-in kept; noise suppressed}
\begin{Verbatim}[commandchars=\\\{\},fontsize=\footnotesize]
\textcolor{diffgray}{@@ news_fda_action @@}
\textcolor{diffredtext}{- IF: FDA approval/rejection/CRL  THEN: signal weight x2.0}
\textcolor{diffgreentext}{+ IF: FDA approval/rejection/CRL  THEN: weight x2.5, directional}
\textcolor{diffgreentext}{+   (approval -> bullish, rejection -> bearish)}
\textcolor{diffgray}{@@ temporal_priced_in @@}
\textcolor{diffredtext}{- IF: moved >3% in news direction  THEN: Reduce signal by 70%}
\textcolor{diffgreentext}{+ IF: FDA+ news AND moved >7%      THEN: Discount signal by 50%}
\textcolor{diffgray}{@@ news_generic_market @@}
\textcolor{diffredtext}{- IF: generic market news  THEN: Reduce weight by 60%}
\textcolor{diffgreentext}{+ IF: generic market news  THEN: Reduce weight by 80%}
\textcolor{diffgray}{@@ news_count_low @@}
\textcolor{diffredtext}{- IF: fewer than 3 articles       THEN: Cap confidence at 0.4}
\textcolor{diffgreentext}{+ IF: <3 significant articles     THEN: Cap confidence at 0.3}
\end{Verbatim}

\vspace{5mm}
\textbf{(b) Consumer Disc.} \hfill {\scriptsize Priced-in narrowed to exception; noise tightened; new symmetric VIX rule}
\begin{Verbatim}[commandchars=\\\{\},fontsize=\footnotesize]
\textcolor{diffgray}{@@ temporal_priced_in @@}
\textcolor{diffredtext}{- IF: moved >3% SAME direction as news   THEN: Reduce signal by 70%}
\textcolor{diffgreentext}{+ IF: moved >3% OPPOSITE to news AND pos sentiment}
\textcolor{diffgreentext}{+                                         THEN: Do NOT discount}
\textcolor{diffgray}{@@ news_generic_market @@}
\textcolor{diffredtext}{- IF: generic market news   THEN: Reduce weight by 60%}
\textcolor{diffgreentext}{+ IF: generic market news   THEN: Reduce weight by 70%}
\textcolor{diffgray}{@@ macro_adjustment_high_fear (new) @@}
\textcolor{diffgreentext}{+ IF: VIX > 25   THEN: Down-weight bullish 30%,}
\textcolor{diffgreentext}{+                      up-weight bearish 30% (symmetric)}
\end{Verbatim}

\caption{Full sector-specific rubric diffs: initial rules (\textcolor{diffredtext}{red}) vs.\ evolved rules (\textcolor{diffgreentext}{green}).}
\label{fig:diffs_app}
\end{figure*}

\section{Free-Form Reflection (A2 Ablation) Example}
\label{app:reflect_example}

\cref{fig:reflect_example} shows the GPT-4.1-mini analyst prompt before and after five rounds of free-form reflection in the A2 ablation (\cref{sec:ablation}).
The v0 prompt is fair-initialized with the same seven rubric rules as the main \method system.
After five rounds, the modular \texttt{IF/THEN} rules have been rewritten into flowing prose whose thresholds quietly drift (e.g., earnings discount $50\% \to 70\%$, VIX damping $30\% \to 40\%$) without any link to specific error days.
Unlike \method's diffs (\cref{fig:diffs}, \cref{app:diffs}), where every threshold change carries a full audit trail back to the error days that motivated it, the free-form variant offers no mechanism to trace or selectively revert individual changes.
This matches \cref{tab:ablation}: without structural constraints, the validation gate can only accept or reject the entire rewritten prompt, and cannot prevent gradual erosion of well-calibrated rules.

\begin{figure}[htbp]
\centering
\begin{Verbatim}[commandchars=\\\{\},fontsize=\scriptsize,frame=single]
\textcolor{diffgray}{@@ analyst_prompt  (v0 fair init)  -->  (v5 after 5 rounds) @@}
\textcolor{diffredtext}{- ...base prompt (JSON schema, general instructions)...}
\textcolor{diffredtext}{- === ANALYSIS RUBRIC (v0, 7 rules) ===}
\textcolor{diffredtext}{- [TEMPORAL_DISCOUNT]}
\textcolor{diffredtext}{-   * IF moved >3% in news direction (5d) THEN reduce signal by 70%}
\textcolor{diffredtext}{-   * IF within 5d of earnings THEN reduce non-earnings news by 50%}
\textcolor{diffredtext}{- [NEWS_WEIGHTING]}
\textcolor{diffredtext}{-   * IF analyst upgrade/downgrade THEN weight x1.3 (top-tier) / x1.0}
\textcolor{diffredtext}{-   * IF generic market commentary THEN reduce weight by 60%}
\textcolor{diffredtext}{-   * IF <3 articles THEN cap confidence at 0.4}
\textcolor{diffredtext}{-   * IF major corporate event THEN signal weight x2.0}
\textcolor{diffredtext}{- [MACRO_INTERACTION]}
\textcolor{diffredtext}{-   * IF VIX > 25 THEN reduce bullish returns by 30%}
\textcolor{diffgreentext}{+ As a senior analyst, your predictions must rigorously integrate price}
\textcolor{diffgreentext}{+ momentum, news catalysts, earnings proximity, and the macro backdrop.}
\textcolor{diffgreentext}{+}
\textcolor{diffgreentext}{+ When a stock has already rallied more than 3% on the same news over}
\textcolor{diffgreentext}{+ the past five days, treat the move as largely priced in and discount}
\textcolor{diffgreentext}{+ the remaining upside aggressively---by at least 70%, or over 90% if}
\textcolor{diffgreentext}{+ the move exceeds 7%.}
\textcolor{diffgreentext}{+}
\textcolor{diffgreentext}{+ Near earnings, be especially skeptical of non-earnings headlines:}
\textcolor{diffgreentext}{+ reduce their weight by roughly 70% and keep confidence low.}
\textcolor{diffgreentext}{+}
\textcolor{diffgreentext}{+ Analyst upgrades from top-tier banks deserve a moderate uplift}
\textcolor{diffgreentext}{+ (around 1.5x), but generic market commentary should be heavily}
\textcolor{diffgreentext}{+ discounted.  Major corporate events (M&A, FDA, earnings) warrant}
\textcolor{diffgreentext}{+ doubling the signal weight.}
\textcolor{diffgreentext}{+}
\textcolor{diffgreentext}{+ In a high-fear regime (VIX above 25), dampen any bullish leanings}
\textcolor{diffgreentext}{+ by about 40% and weigh geopolitical tail risks more seriously.}
\textcolor{diffgreentext}{+}
\textcolor{diffgreentext}{+ When signals conflict, err toward lower confidence rather than}
\textcolor{diffgreentext}{+ forcing a directional call.  Keep reasoning concise, grounded in}
\textcolor{diffgreentext}{+ recent price action, and focused on material events from the past}
\textcolor{diffgreentext}{+ three to five days.}
\end{Verbatim}
\caption{Free-form reflection (A2) on AI~Tech, GPT-4.1-mini, window~0.
The v0 prompt is fair-initialized with the same rubric content as the main \method system.
Without structural constraints, the writing style gradually shifts from modular \texttt{IF/THEN} rules into flowing prose, and individual changes become difficult to track across rounds.}
\label{fig:reflect_example}
\end{figure}

\section{Attribution Agent Example Trace}
\label{app:attribution}

\cref{fig:attribution_example_app} provides a compact textual trace of the attribution process: the agent receives the worst-performing days, diagnoses a recurring error pattern, and proposes a rule fix.

\begin{figure}[htbp]
\centering
\small
\textbf{Step 1.} Backtest on training period (2025-04-15 to 2025-08-14) finds 20 worst portfolio days. For each, pack full context:
\begin{Verbatim}[fontsize=\scriptsize,frame=single]
Day 7: 2025-05-16 (portfolio return: -2.31%)
  Long: [NVDA, GOOGL, CRM, AMZN, META]  Short: [AAPL, PLTR, AI, SNOW, ARM]
  [NVDA] predicted: +5.0% (bullish, conf=0.75)
    reason: "strong AI chip demand; positive trade news"
    rules_applied: [news_high_impact, macro_low_vix]
  Actual NVDA return: -2.8%  (5d prior: +16.1%)
  Headlines: "Nvidia stock set for 15% weekly gain as trade
             news boosts AI chip"
\end{Verbatim}

\vspace{1mm}
\textbf{Step 2.} LLM analyzes all 20 days jointly $\to$ 5 structured error patterns:
\begin{Verbatim}[fontsize=\scriptsize,frame=single]
[insufficient_temporal_discount_for_recent_momentum] freq=17
  "The analysis process often predicts bullish returns
   despite stocks having already rallied strongly on the
   same news. temporal_priced_in threshold too loose."

[overreliance_on_analyst_upgrades_and_price_targets] freq=15
  "High weight on analyst rating changes and price target
   upgrades without adjusting for recent price run-up."

[insufficient_downweighting_of_news_near_earnings] freq=10
  "Non-earnings news near earnings dates not sufficiently
   downweighted; LLM overreacts to noise pre-earnings."
\end{Verbatim}

\vspace{1mm}
\textbf{Step 3.} Evolution agent proposes 3 atomic edits targeting the top patterns:
\begin{Verbatim}[fontsize=\scriptsize,frame=single]
[modify] temporal_priced_in:
    Increase discount for news already reflected in price
[modify] news_analyst_rating:
    Reduce weight and cap confidence for analyst upgrades
[modify] temporal_earnings_season:
    Increase downweighting of non-earnings news near earnings
\end{Verbatim}
Rubric evolved: v0 $\to$ v1 (3 changes applied, accepted on validation).
\caption{Illustrative example of the attribution agent pipeline.}
\label{fig:attribution_example_app}
\end{figure}

\newpage
\clearpage

\end{document}